\pdfoutput=1

\documentclass[11pt]{article}
\usepackage[preprint]{acl}
\usepackage{times}
\usepackage{latexsym}
\usepackage{amsmath}
\usepackage{amssymb}
\usepackage{booktabs}
\usepackage{graphicx}
\usepackage{multirow}
\usepackage{xcolor}
\usepackage{url}

\title{When Does a Second Model Help?\\Cross-Model Review in LLM Verification}

\author{Song Tae-Eun \\
  Daejeon Jungang Cheonggua Co., Ltd. \\
  \texttt{higheun@gmail.com}}

\begin{document}
\maketitle

\begin{abstract}
Large language models now generate code, documentation, and analyses, and are increasingly used to review such output. We ask when a second review by a different model helps. Building on the author's earlier preprints, which varied context, repetition, and role structure within one model, we test model independence in a controlled experiment: 30 artifacts with 150 planted errors, 10 review conditions, and 900 review sessions with three reviewer models from two developers. In this experiment, (1) a top-tier cross-model reviewer is not significantly different in F1 from same-model review in a fresh session (CCR), which does not establish equivalence; (2) the two find partly different errors (Jaccard 41.2\%); and (3) at two review calls, one CCR plus one cross-model review matches more planted errors than two CCR reviews (56.7\% vs.\ 42.7\%; Holm-adjusted $p=.006$), but not significantly more than two reviews by the top-tier cross-model reviewer, so model difference and reviewer capability are not separated. A lightweight cross-model reviewer scores no higher than same-model review. Withholding requirements from the reviewer raises F1 for the two lower tiers but not the top tier, in untested point estimates whose pattern depends on how failed sessions are scored. Before analysis we audited all session records, excluding one baseline run of uncertain provenance and 14 failed calls; results with all sessions are also reported. A partial check on public detector outputs from another benchmark neither replicates nor contradicts the main comparison. Records, artifacts, and scripts are available from the author on request.
\end{abstract}

\section{Introduction}

Large language models (LLMs) now generate code, documentation, data analyses, and research artifacts faster than people can review them. This has motivated the use of LLMs themselves as reviewers, an approach that is already gaining industrial traction, as evidenced by the recent release of cross-model review tooling integrated directly into AI coding assistants \citep{huet2026crossmodel}. Two questions are open: what makes AI review of AI output independent, and does cross-model review improve defect detection?

Verification is only as useful as it is independent of what produced the output. In human contexts, peer review, double-blind evaluation, and independent auditing all derive their value from ensuring that the verifier's judgment is not contaminated by the same biases that produced the original output. When the same LLM reviews its own generation within the same conversational context, it shares its training data, parameters, and conversational context, and in-context self-correction without external feedback has been found to degrade accuracy \citep{huang2024large}.

A series of earlier studies by the author explored individual dimensions of independence within one model. Cross-Context Review \citep[CCR;][]{author2026a} reported that \textit{context independence} (removing the generation conversation before review) improves defect detection; in the data reused here, after excluding one baseline run of uncertain provenance, CCR exceeds same-context self-review by only 1.5pp in F1, which is not significant (Section~\ref{sec:overall}); the two also differ in whether the reviewer sees the original request. Diachronic CCR \citep[D-CCR;][]{author2026b} reported that adding repeated review rounds, which removes \textit{temporal independence} in our terms, is counterproductive, increasing false positives by 62\% without improving recall. In a study whose main subject is benchmark-contamination detection, the Hierarchical Cross-Context Architecture report \citep[HCCA;][]{author2026c} also describes a pilot in which a multi-stage pipeline (worker, verifier, director) whose later stages could see earlier conclusions produced only sycophantic confirmation, and identify \textit{information restriction} rather than structural complexity as the key mechanism. These three reports are the author's own single-author preprints that have not been peer reviewed, so we do not rely on their conclusions: every baseline number below is recomputed from the per-session data, the benchmark and matching procedure are specified in Appendix~\ref{app:benchmark}, and where our test disagrees with an earlier report we say so.

All three studies stayed within one model. The most intuitive form of independence, using a \textit{different} model, remained untested. This paper addresses that gap.

We introduce a four-dimensional framework for independence in LLM verification, encompassing context, temporal, structural, and model independence. Through a controlled experiment with 900 review sessions across 10 conditions, we systematically evaluate model independence and its interaction with other dimensions. Our key findings are:

We use \emph{complementary} to mean partially non-overlapping detection, not super-additive coverage. (1)~\textbf{Complementary detection}: cross-model review (XMR) and CCR find partly different defects (Jaccard 41.2\%), with CCR ahead on code (F1 40.7\%) and XMR on documents (F1 up to 42.1\%); (2)~\textbf{call-matched coverage}: one CCR plus one XMR review matches 56.7\% of planted errors against 42.7\% for two CCR reviews, while the gain over two top-tier XMR reviews is not significant; (3)~\textbf{reviewer capability appears to shape the effect}---a top-tier reviewer shows no significant difference from CCR, but a lightweight model scores numerically below even baseline same-context review, so the value of model independence appears to depend on reviewer quality; and (4)~\textbf{information restriction} raises cross-model F1 for the two lower tiers (+2.2 and +3.0pp) but not the top tier ($-0.2$pp; untested point estimates).

These results suggest tentative guidance: pairing a same-model review with a cross-model one gave more coverage than two same-model reviews, though not significantly more than two top-tier cross-model reviews; for resource-constrained settings, a single cross-model review with a top-tier reviewer had the highest F1 of any single condition in our data (not significantly above CCR); and lightweight cross-model reviewers offered no advantage over same-model review.

\section{Background}

\subsection{Context Independence}

Cross-Context Review \citep[CCR;][]{author2026a} conducts review in a fresh session, so the reviewer receives only the artifact, not the conversation that produced it. Across 360 sessions over 30 artifacts with 150 ground truth errors, the original report found CCR ahead of same-context review, testing run~1 alone (paired $t$, $p=0.008$); among the four same-model conditions, CCR ranks first in that run only (run-level F1 in Appendix~\ref{app:itt}). In the data used here, after excluding one same-context run of uncertain provenance (Section~\ref{sec:provenance}), CCR's F1 is 28.6\% versus 27.1\% (a comparison that changes both the context and whether the request is visible; Section~\ref{sec:framework-conds}), and the artifact-level test finds no significant difference (unadjusted $p=0.26$; Table~\ref{tab:main_results}). The original report attributed the gap to removing the conversational anchoring that biases a model toward defending its own output.

\subsection{Temporal Independence}

Diachronic CCR \citep[D-CCR;][]{author2026b} tested repeated review rounds: counter-intuitively, a second round \textit{decreased} performance, with false positives increasing by 62\%. The report attributes this to the same model amplifying its existing false-confidence patterns rather than gaining new perspectives, which would suggest that repeating review in the same session is not a path to independence.

\subsection{Structural Independence}

The HCCA report \citep{author2026c} assigns specialized analytical roles under intentional information restriction. In a pilot extending this to a multi-stage pipeline (worker, verifier, director) in which later stages could see earlier conclusions, the report observes only sycophantic confirmation and attributes the benefit of its main design to information restriction rather than to structural complexity.

\subsection{The Remaining Dimension: Model Independence}

The three prior studies explored three ways of varying review \textit{within a single model}: context can be reset, timing varied, and roles restructured, but the model's training data, weights, and systematic biases remain constant. The natural next question is whether a \textit{different} model, trained on different data with different objectives, provides a fundamentally different form of independence.

Indirect evidence suggests it should. LLMs differ in measured capability across tasks \citep{zheng2023judging,chiang2024chatbot}, and cross-model review has already been integrated into production developer tools \citep{huet2026crossmodel}. However, we are not aware of prior work that systematically measures whether these differences translate into complementary defect detection.

\subsection{Related Work}

\textbf{LLM self-correction and self-preference.} In-context self-correction without external feedback degrades accuracy \citep{huang2024large}; \citet{kamoi2024when} attribute the failure to the feedback (error-detection) stage and recommend analysing the quality of generated feedback directly. They also note that cross-model correction is unsuitable for testing whether a model can improve its own responses, and recommend comparisons against baselines of comparable computational cost; our two-review-call analysis matches the number of review calls, not compute. LLMs also favor their own generations \citep{panickssery2024llm}. Neither line compares a same-conversation reviewer against a fresh-context or different-model reviewer: the axis we vary.

\textbf{Sycophancy and evaluation bias.} LLMs adjust outputs to match perceived user preferences \citep{sharma2024sycophancy} and, as judges, exhibit positional and other judgement biases \citep{wang2024notfair,chen2024humans}.

\textbf{Multi-agent systems and LLM-as-judge.} Multi-agent frameworks structure multiple LLM instances to debate answers or assume specialized roles \citep{du2024improving,liang2024encouraging,chan2024chateval,hong2024metagpt,qian2024chatdev}, while LLM-as-judge approaches use LLMs to score or rank outputs \citep{zheng2023judging,li2024llmsjudges,kim2024prometheus}. These systems primarily manipulate structural and context independence but typically use a single underlying model, leaving model independence unexplored. Error-detection benchmarks: ReaLMistake \citep{kamoi2024evaluating} and BIG-Bench Mistake \citep{tyen2024llms} report low detection rates even for strong models, and CriticBench \citep{lin2024criticbench} finds that stronger models critique weaker ones better while weaker models can sometimes surpass stronger ones in self-critique; but none of these analyses is organised around a model inspecting its own output versus another model's.

\textbf{Software review and blind spots.} LLMs are increasingly applied to code review \citep{li2022codereviewer,pornprasit2024finetuning,cihan2025automated}; in a related setting, \citet{shin2025blind} find that LLM-generated reviews of research papers over-focus on technical validity and overlook novelty compared with human experts. \citet{doddapaneni2024finding} probe five evaluator LLMs with 2,400 perturbed answers and report that they miss quality drops in over 50\% of cases on average. We use ``blind spot'' in a narrower sense (defects one reviewer model misses and another catches) and quantify it through Jaccard overlap.

\textbf{Hallucination detection, ensembles, and N-version.} Consistency-based methods exploit the intuition that disagreement across independent samples signals unreliability, via semantic entropy \citep{farquhar2024detecting} or reference checking \citep{manakul2023selfcheckgpt}. Ensemble and mixture-of-agents methods show that model diversity improves aggregate performance \citep{dietterich2000ensemble,wang2024moa,tekin2024llmtopla}, echoing N-version programming \citep{chen1978nversion}; we extend these principles to across-model defect detection.

\textbf{Correlated errors and frontier blind spots.} \citet{kim2025correlated} analyze over 350 LLMs, largely on multiple-choice benchmarks; on one leaderboard (HELM), pairs of models that both err choose the same wrong answer about 60\% of the time, and larger, more accurate models have more highly correlated errors. This result raises concerns about naive multi-model ensembles and motivates our measurement of pairwise finding overlap (52.2\% within the same model vs.\ 37--46\% across models; Section~\ref{sec:h3}), a descriptive comparison rather than a test of error correlation.

\section{A Framework for Independence in LLM Verification}

\subsection{Dimensions of Independence}

We define four binary dimensions characterizing the relationship between a generator LLM (which produces the artifact) and a reviewer LLM (which verifies it):

\textbf{Context independence} (C): the reviewer has no access to the generation conversation, operating with $P(\text{findings} \mid \text{artifact})$ rather than $P(\text{findings} \mid \text{artifact}, G)$, where $G$ is the generation transcript.

\textbf{Temporal independence} (T): the review occurs in a single pass, with no memory of prior review rounds on the same artifact.

\textbf{Structural independence} (S): the reviewer operates without hierarchical subordination to the generator, avoiding deference or sycophancy.

\textbf{Model independence} (M): the reviewer is a different model, typically from a different developer. This does not guarantee different training data or uncorrelated errors; whether cross-model errors are less correlated is an empirical question (Section~\ref{sec:h3}).

We additionally define information restriction (R) as a cross-cutting modifier: the reviewer receives only the artifact, without the original prompt or requirements, forcing assessment on intrinsic merits rather than compliance with stated goals.

\subsection{Experimental Conditions as Dimension Combinations}\label{sec:framework-conds}

Our 10 conditions map onto the framework as shown in Table~\ref{tab:conditions}. All XMR conditions inherently include context independence, because in our design the cross-model reviewer never receives the generation conversation. In the experiment code, CCR is specified with the same user-prompt string as the information-restricted XMR conditions (the artifact only), and SA with the same structure as the unrestricted ones (the original request and the artifact). In terms of the specified prompt, the comparisons closest to changing only the reviewer model are therefore CCR versus the -R conditions and SA versus the unrestricted XMR conditions (the latter also changes structure if SA's sub-agent role matters); comparing CCR with an unrestricted XMR condition changes both the reviewer model and the information given, and comparing CCR with SR changes both the context and the information given. The reviewers also run under different system instructions (the Claude Code agent tool, the Codex CLI agent, or none for the Gemini API), and replacing the reviewer model changes reviewer capability and other model characteristics, so none of these comparisons isolates model independence.

\begin{table*}[t]
\centering
\small
\begin{tabular}{lcccll}
\toprule
\textbf{Condition} & \textbf{C} & \textbf{M} & \textbf{R} & \textbf{Reviewer Model} & \textbf{Description} \\
\midrule
SR       & $-$ & $-$ & $-$ & Generator (same session) & Baseline: self-review \\
SR2      & $-$ & $-$ & $-$ & Generator (2nd pass)     & Repeated review (T=0) \\
SA       & $+$ & $-$ & $-$ & Generator (sub-agent)    & New session with request (S varied) \\
CCR      & $+$ & $-$ & $+$ & Generator (new session)  & New session, artifact only \\
\midrule
XMR-Ge     & $+$ & $+$ & $-$ & Gemini 2.5 Flash  & Cross-model (lightweight) \\
XMR-Ge-R   & $+$ & $+$ & $+$ & Gemini 2.5 Flash  & + information restriction \\
XMR-GePro  & $+$ & $+$ & $-$ & Gemini 2.5 Pro        & Cross-model (capable) \\
XMR-GePro-R& $+$ & $+$ & $+$ & Gemini 2.5 Pro        & + information restriction \\
XMR-GPT    & $+$ & $+$ & $-$ & GPT-5.4        & Cross-model (top-tier) \\
XMR-GPT-R  & $+$ & $+$ & $+$ & GPT-5.4        & + information restriction \\
\bottomrule
\end{tabular}
\caption{Experimental conditions mapped to independence dimensions. Temporal (T) and structural (S) dimensions are held constant across most conditions; SR2 uniquely lacks temporal independence (repeated review), and SA uniquely varies structural independence (sub-agent role). See Sections 2.2--2.3 for detailed analysis of these dimensions. SA receives the original request with the artifact, as the unrestricted XMR conditions do, and CCR receives the artifact only, as the -R conditions do (prompts in Appendix~\ref{app:prompts}). Version~1 of this paper marked CCR as unrestricted and SA as lacking context independence; both labels were inconsistent with the experiment code.}
\label{tab:conditions}
\end{table*}

\subsection{Hypotheses}

Based on our framework, we test the following hypotheses:

\textbf{H1 (Unique Finding).} Cross-model review discovers defects that cross-context review misses, and vice versa. Formally: $\text{UFR}(\text{XMR} \to \text{CCR}) > 0$ and $\text{UFR}(\text{CCR} \to \text{XMR}) > 0$, where $\text{UFR}(A \to B)$ is the proportion of $A$'s true positives not found by $B$.

\textbf{H2 (Union Superiority).} The union of CCR and XMR findings achieves higher recall than either alone (implied by H1 under a fixed matcher; we report it for its magnitude). Formally: $\text{Recall}(\text{CCR} \cup \text{XMR}) > \max(\text{Recall}(\text{CCR}), \text{Recall}(\text{XMR}))$.

\textbf{H3 (Bias Orthogonality).} The overlap between CCR and XMR findings is lower than the overlap between independent sessions within the same model, which would be consistent with systematically different blind spots. Formally: $\text{Jaccard}(\text{CCR}, \text{XMR}) < \text{Jaccard}(\text{CCR}_A, \text{CCR}_B)$.

We additionally explore two non-directional hypotheses:

\textbf{H4 (Dimension Interaction).} The combination of context and model independence exhibits interaction effects (synergistic or antagonistic) beyond the sum of individual effects. This hypothesis is not identifiable in our design, which has no condition with model independence but no context independence; we state it for completeness.

\textbf{H5 (Error-Type Specialization).} Different independence dimensions are differentially effective across error types (FACT, CONS, CTXT, RCVR, MISS) and artifact categories (code, document, script).

\section{Experimental Setup}

\subsection{Experiment A: Controlled Code Review}

\textbf{Artifacts.} We reuse the 30 software artifacts and 150 ground truth errors from \citet{author2026a}. The artifacts span three categories: code modules (C1--C10), technical tutorials (D1--D10), and presentation scripts (S1--S10). Each artifact contains 5 planted defects across 5 error types: factual errors (FACT), internal consistency violations (CONS), missing context handling (CTXT), recoverability gaps (RCVR), and missing functionality (MISS). Error severities range from Critical to Minor. Reusing identical artifacts enables direct comparison with prior results.

\textbf{Generator model.} All 30 artifacts were generated by a single frontier-class LLM, Claude Opus 4.6 according to our run logs; the artifact files themselves do not record the generating model. The same-model baselines (SR, SR2, SA, CCR) were run with that model, subject to the same caveat. Reviewer model identifiers denote the requested model setting (an API parameter for Gemini, a Codex CLI setting for GPT-5.4) rather than a confirmed served checkpoint; we did not record a served model identifier. Using a single generator ensures that all reviewer conditions evaluate identical artifacts, removing generator variation; it does not separate reviewer independence from other reviewer characteristics such as capability.

\textbf{Dependence on prior work.} The four baseline conditions (SR, SR2, SA, CCR), the C/T/S dimensions, and the artifacts with their ground truth come from the earlier preprints; only the XMR conditions and the unifying taxonomy are new. Appendix~\ref{app:benchmark} specifies the artifacts, error injection, and matching so that the claims here can be checked without those reports. The baselines were collected before the XMR conditions under the same protocol, a temporal asymmetry noted in Limitations.

\textbf{Data provenance and exclusions.}\label{sec:provenance} We audited all 900 session records against the raw files before the analysis reported here. The 540 cross-model records carry the reviewer model identifier, call duration, and a per-session timestamp. The 360 same-model baseline sessions were run through the Claude Code agent tool and transcribed into structured files by the host session; they carry no raw model output, and in several runs all 30 files share one timestamp. For one run (SR, run~3), 92\% of the finding descriptions appear verbatim inside the script that wrote the files, so we cannot confirm that they are unedited model output, and we exclude this run from all analyses. Fourteen cross-model sessions failed (five timeouts or crashes and nine unparseable outputs; eight of the fourteen were XMR-Ge-R sessions) and had been scored as reviews with no findings; because a failed call is not a review, we exclude them as well. The analyses therefore describe cross-model review when the call succeeds; Appendix~\ref{app:itt} reports the main values with failures scored as zero. The exclusion rules followed from the audit and were fixed before the full recomputation, although a partial check of their effect on cross-model F1 was run during the audit. The remaining baseline records are unverified rather than known to be clean.

\textbf{Conditions and scale.} We evaluate 10 conditions (Table~\ref{tab:conditions}), each with 90 sessions (3 runs per artifact $\times$ 30 artifacts), yielding 900 sessions total.

\textbf{Reviewer models.} We employ three external reviewer models from two developers at different capability tiers. The tiers reflect the vendors' positioning at the time of the experiment, not a capability we measured, and GPT-5.4 is also a later model generation than Gemini~2.5:
\begin{itemize}
    \item \textit{GPT-5.4} (top-tier): a model from a different developer than the generator, run through the Codex CLI agent (see Limitations on its working directory).
    \item \textit{Gemini 2.5 Pro} (capable): a strong model from a developer other than those of the generator and GPT-5.4.
    \item \textit{Gemini 2.5 Flash} (lightweight): a cost-optimized model from the same developer as Gemini 2.5 Pro, testing whether model independence helps even with reduced capability.
\end{itemize}

For each family, we test both unrestricted (full artifact + requirements) and information-restricted (artifact only) conditions.

\textbf{Procedure.} Each session proceeds identically: the reviewer receives the artifact (in SA and the unrestricted XMR conditions, also the original request, i.e., the requirements; SR and SR2 review in the generation session; prompts in Appendix~\ref{app:prompts}) and produces a structured list of findings (location, description, severity, error type). Findings are matched against the ground truth by location- and keyword-overlap scoring, each finding assigned to at most one ground-truth error and each ground-truth error credited at most once per review (Appendix~\ref{app:benchmark}).

\textbf{Statistical testing.} We use the Wilcoxon signed-rank test for paired comparisons at the artifact level (averaging the available runs per artifact per condition after the exclusions of Section~\ref{sec:provenance}; every artifact keeps at least one run in every condition, so each test uses 30 paired observations), with Holm-Bonferroni correction over the nine comparisons against CCR. For comparison with the original CCR report, Table~\ref{tab:main_results} also gives paired $t$-tests on run~1 alone without correction; our conclusions use the former. Union analysis computes the unique ground truth errors detected by at least one session per condition, yielding Union Recall, Unique Finding Rate (UFR), and Jaccard similarity.

\subsection{Reproducibility}

The artifacts, ground-truth annotations, review prompts, per-session review records, and analysis scripts are available from the author on request. Raw reviewer outputs and per-session model identifiers exist only for the cross-model conditions; the same-model baseline records were transcribed into structured files by the host session and do not include raw model output (Section~\ref{sec:provenance}). We note that exact reproduction is limited by the non-deterministic nature of LLM outputs, and the 90 sessions per condition are repeated measurements on 30 artifacts, so our tests use 30 artifact-level pairs.

\section{Results}

\subsection{Overall Performance}\label{sec:overall}

\begin{figure}[t]
\centering
\includegraphics[width=\columnwidth]{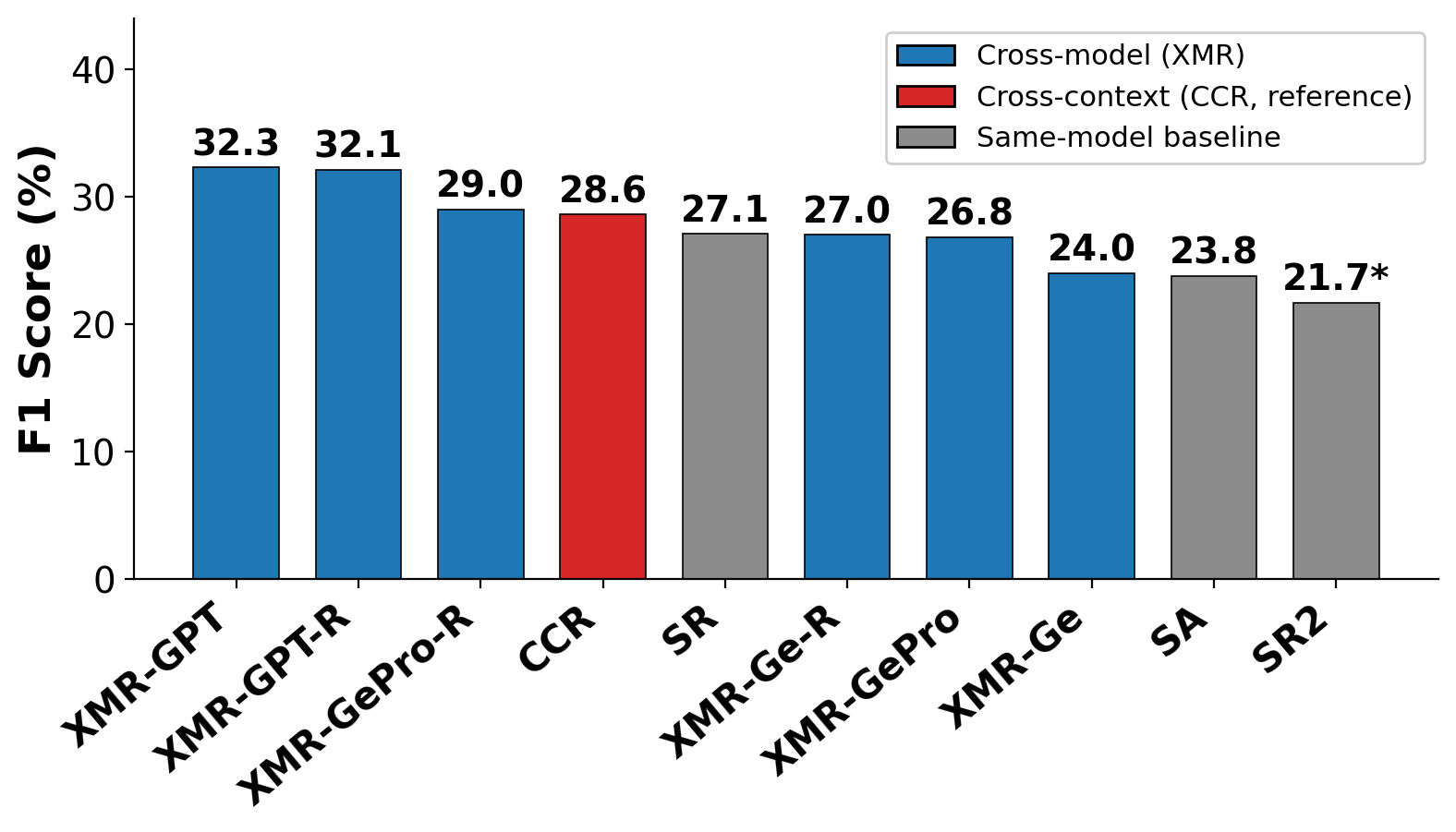}
\caption{F1 scores for all 10 conditions (ranked). The asterisk marks the only condition significantly different from CCR (SR2; Wilcoxon signed-rank on per-artifact means over the available runs, Holm-adjusted $p=0.011$); full statistics in Table~\ref{tab:main_results}.}
\label{fig:f1_bar}
\end{figure}

Table~\ref{tab:main_results} presents the main results across all 10 conditions, ranked by F1 score, and Figure~\ref{fig:f1_bar} (Appendix) visualizes this ranking.

\begin{table}[t]
\centering
\small
\setlength{\tabcolsep}{2.6pt}
\begin{tabular}{clcccccc}
\toprule
\textbf{Rank} & \textbf{Condition} & $n$ & \textbf{F1} & \textbf{Prec.} & \textbf{Rec.} & \textbf{$p_{\mathrm{W}}$} & \textbf{$p_t$} \\
\midrule
1 & XMR-GPT     & 86 & 32.3 & 28.6 & 38.4 & 1.00 & .478 \\
2 & XMR-GPT-R   & 89 & 32.1 & 27.5 & 39.8 & 1.00 & .524 \\
3 & XMR-GePro-R & 90 & 29.0 & 29.2 & 29.3 & 1.00 & .176 \\
4 & CCR         & 90 & 28.6 & 31.5 & 27.1 & ---  & ---  \\
5 & SR          & 60 & 27.1 & 26.2 & 28.3 & 1.00 & .008 \\
6 & XMR-Ge-R    & 82 & 27.0 & 34.2 & 23.9 & 1.00 & .232 \\
7 & XMR-GePro   & 90 & 26.8 & 30.6 & 24.9 & 1.00 & .083 \\
8 & XMR-Ge      & 89 & 24.0 & 35.1 & 20.0 & 1.00 & .017 \\
9 & SA          & 90 & 23.8 & 27.4 & 21.8 & .228 & .004 \\
10& SR2         & 90 & 21.7 & 21.0 & 22.7 & \textbf{.011} & $<$.001 \\
\bottomrule
\end{tabular}
\caption{Performance across 10 conditions (900 sessions; $n$ = sessions analysed after excluding SR run~3 and 14 failed cross-model sessions, Section~\ref{sec:provenance}; Appendix~\ref{app:itt} gives the values with failures scored as zero). Values are percentages. $p_{\mathrm{W}}$: Wilcoxon signed-rank test against CCR on per-artifact means over the available runs, Holm-adjusted over nine comparisons (only SR2 differs); $p_t$: paired $t$-test on run~1 only, unadjusted, as in the original CCR report.}
\label{tab:main_results}
\end{table}

The top three conditions are all cross-model configurations, and the bottom three are the lightweight cross-model condition (XMR-Ge), the sub-agent baseline (SA), and the repeated baseline (SR2). XMR-Ge-R (27.0\%) is level with SR (27.1\%), so model independence with a lightweight reviewer shows no clear advantage over baseline self-review.

The difference between the best XMR condition (XMR-GPT, 32.3\%) and CCR (28.6\%) is \textit{not} statistically significant (adjusted $p_{\mathrm{W}}=1.00$; run-1 $p_t=0.478$). We therefore detect no F1 difference between these two conditions; without an equivalence test, this is not evidence that the two are equal. XMR-GPT also receives the original request, which CCR does not; the comparison with the same prompt as CCR, XMR-GPT-R (32.1\%), is likewise not significant (adjusted $p_{\mathrm{W}}=1.00$).

The spread across reviewer models is larger. XMR-GPT (32.3\%) exceeds XMR-Ge (24.0\%) by 8.3 percentage points, more than the gap between CCR and SR (1.5pp). This single comparison is outside the nine-comparison family; its unadjusted Wilcoxon $p$ is 0.020 with failed sessions excluded and 0.071 with them scored as zero, and it would not survive correction if added to the family. The data are consistent with reviewer capability mattering more than independence per se, but do not establish it.

\subsection{Complementary Detection (H1, H2)}\label{sec:union}

While individual F1 scores are similar between CCR and top XMR conditions, their \textit{coverage patterns} differ meaningfully.

\begin{figure}[t]
\centering
\includegraphics[width=\columnwidth]{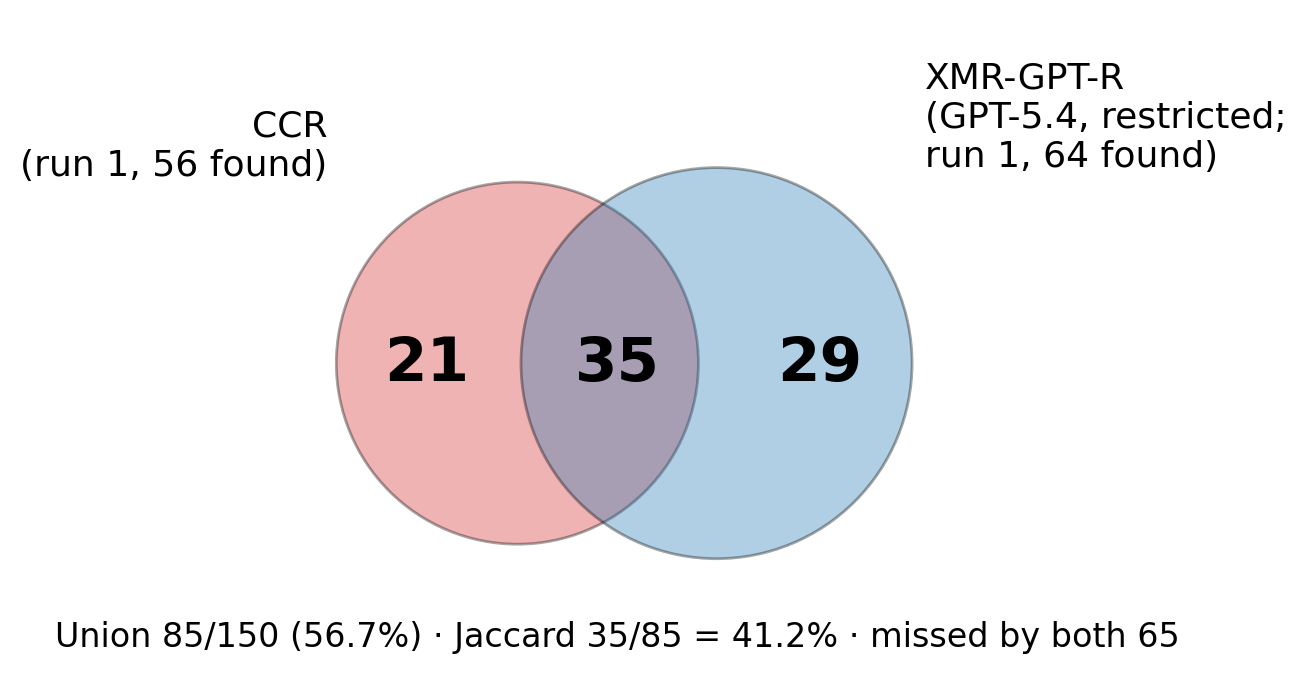}
\caption{Planted errors matched by CCR and by XMR-GPT-R (run~1 of each condition; 150 errors in total).}
\label{fig:venn}
\end{figure}

\textbf{Union Recall} (Figure~\ref{fig:venn}): under the existing automated matcher, a two-review-call comparison yields 85/150 (56.7\%) matched ground-truth errors for one CCR plus one XMR-GPT-R review versus 64/150 (42.7\%) for two CCR reviews (all four reviews are specified with the same artifact-only prompt), a 14.0pp gain (artifact-level bootstrap 95\% CI [6.7, 22.0]; signed-rank $p=.006$, Holm-adjusted over the three union comparisons). These are set-level recalls over run~1 (singles 56 and 64; $56 + 64 - 35 = 85$), not the per-session means of Table~\ref{tab:main_results}. Across the three runs per condition, mixed unions cover 44.7--56.7\% and CCR-only unions 30.0--42.7\% (reused runs, not independent replications). Against two XMR-GPT-R reviews, on the 29 artifacts without a failed session (76/145 vs.\ 82/145 for the mixed pair), the gain is 4.1pp (95\% CI [$-2.1$, 10.3]; adjusted $p=.184$). The third comparison in the family uses self-review as the same-model partner: SR plus XMR-GPT-R matches 77/150 against 53/150 for SR plus SR2 in run~1 (+16.0pp, 95\% CI [6.7, 26.0]; adjusted $p=.008$). These retrospective results are conditional on the matcher; condition-specific matching bias remains unmeasured.

The Jaccard similarity between CCR and XMR-GPT-R is 41.2\% (35 of 85 unique errors found by either), indicating moderate overlap: the approaches share a substantial core, but $\sim$59\% of unique findings come from only one approach, supporting H1 (each finds defects the other misses). The 14.0pp union gain is consistent with H2 (union superiority) in the narrow sense that the union covers more than either review alone. Any two imperfect, non-identical detectors raise union recall mechanically; the observed unions in fact fall below both pooled and within-artifact independence reference values ($r_A + r_B - r_A r_B$). These descriptive comparisons do not establish synergy or its absence; we do not claim synergy (super-additive coverage beyond independent expectation)---complementarity here refers to non-overlap (Jaccard 41.2\%).

\subsection{Bias Orthogonality (H3)}\label{sec:h3}

To test whether cross-model overlap is lower than within-model overlap (which would be consistent with systematically different blind spots), we computed pairwise Jaccard similarity at the finding level across all run pairs.

\textbf{Within-condition overlap} (same model, different sessions): CCR 52.2\% (90 run pairs), SR 61.8\% (30 pairs, runs 1--2 only), XMR-GPT 59.3\% (84), XMR-Ge 55.1\% (88). Cross-condition overlap (all run combinations): CCR vs.\ XMR-GPT 46.1\% (258 pairs), CCR vs.\ XMR-Ge 36.5\% (267), and CCR vs.\ SR 59.9\% (180; same model, different context and information). Pairs involving an excluded session are dropped, so the pair counts differ.

Within-model session pairs share 52--62\% of findings, while cross-model pairs share 37--46\%. The difference between CCR intra-condition overlap (52.2\%) and CCR-vs-XMR cross-model overlap (average 41.3\%) is 10.9 percentage points, consistent with H3. Lower overlap can also arise from differences in detection rates or finding counts, and the run pairs share artifacts, so this is descriptive support rather than a test.

CCR vs.\ SR overlap (59.9\%) is comparable to within-condition overlaps: we see no clear sign that changing the context (together with withholding the request) changes which errors the model finds, whereas changing the model is accompanied by lower overlap, which matters given that high-capability LLMs exhibit strongly correlated errors \citep{kim2025correlated}.

\subsection{Domain Specialization (H5)}

The aggregate F1 scores mask a reversal at the category level (Table~\ref{tab:category_f1}).

\begin{table}[t]
\centering
\small
\begin{tabular}{lcccc}
\toprule
\textbf{Category} & \textbf{CCR} & \textbf{GePro-R} & \textbf{GPT} & \textbf{GPT-R} \\
\midrule
Code     & \textbf{40.7} & 23.4 & 37.2 & 29.8 \\
Document & 24.5 & 38.1 & \textbf{42.1} & 40.7 \\
Script   & 20.7 & 25.6 & 15.4 & \textbf{25.8} \\
\bottomrule
\end{tabular}
\caption{F1 (\%) by artifact category. Bold indicates best condition per category.}
\label{tab:category_f1}
\end{table}

For code artifacts, CCR outperforms all XMR conditions; one possible explanation is that the generating model shares an understanding of programming patterns and code structure that an external model lacks. For documents, the pattern reverses, and it holds for every condition, not only those in the table: all six cross-model conditions (F1 30.1--42.1\%) score above all four same-model conditions (17.4--24.5\%). XMR conditions perform better, suggesting a different model brings different expectations about consistency and completeness, shaped by different training corpora. This category reversal is our clearest descriptive support for H5, although the category estimates carry no uncertainty intervals and the explanations above are conjectures; it suggests that the choice between CCR and XMR should depend on artifact type.

\subsection{Error Type Analysis}

Breaking down results by error type reveals further specialization (Table~\ref{tab:error_type}). Detection rates are per-session recall for each error type, averaged over the analysed sessions ($n$ as in Table~\ref{tab:main_results}).

\begin{table}[t]
\centering
\small
\begin{tabular}{lcccc}
\toprule
\textbf{Error Type} & \textbf{CCR} & \textbf{XMR-Ge} & \textbf{GePro} & \textbf{GPT} \\
\midrule
FACT & 48 & 35 & 48 & \textbf{58} \\
CONS & 37 & 33 & 38 & \textbf{53} \\
CTXT & 13 &  9 &  9 & \textbf{26} \\
RCVR & 19 &  7 & 16 & \textbf{34} \\
MISS & 19 & 17 & 14 & \textbf{21} \\
\bottomrule
\end{tabular}
\caption{Detection rate (\%) by error type (recall per type). Bold indicates best condition per type.}
\label{tab:error_type}
\end{table}

Several patterns emerge. The top-tier cross-model reviewer is higher than CCR on FACT (58\% vs.\ 48\%) and CONS (53\% vs.\ 37\%). On CTXT, where operational context matters, the two top-tier conditions reach 26--31\%, against 9--18\% for every other condition (XMR-GPT-R and SR are not shown in the table). Across all six cross-model conditions (including three not shown in Table~\ref{tab:error_type}), RCVR varies most (7--38\%), followed by FACT (35--58\%) and CTXT (9--31\%), consistent with these types being sensitive to reviewer capability.

The apparent tension between Tables~\ref{tab:category_f1} and \ref{tab:error_type}, namely CCR's higher F1 on code (40.7\% vs.\ 37.2\%) despite XMR-GPT's comparable or higher per-type recall, is consistent with precision: on code, CCR's precision is 46.9\% versus 35.6\% for XMR-GPT, while its recall is 36.7\% versus 40.0\%. The analyses also operate at different granularities, so dominance at one level need not imply it at the other.

\subsection{Information Restriction Effect}\label{sec:restrict}

Restricting the information available to the reviewer, by providing only the artifact without the original requirements, changes cross-model F1 by tier as follows: top-tier 32.3$\to$32.1 ($-0.2$), capable 26.8$\to$29.0 (+2.2), and lightweight 24.0$\to$27.0 (+3.0). The two sides of each comparison are not the same session sets after exclusions, and with failed sessions scored as zero the changes are +0.9, +2.2, and +0.9 (Appendix~\ref{app:itt}).

This pattern is consistent with the earlier HCCA report \citep{author2026c} that less information promotes more independent judgment; the per-tier gains are untested point estimates, and the account that follows is a hypothesis. When the reviewer cannot simply check the artifact against stated requirements, it must assess the artifact on its intrinsic quality, detecting inconsistencies, logical gaps, and errors that requirement-checking might overlook. After the exclusions the top-tier reviewer shows no gain (+0.9 with failures scored as zero), and the size of the lightweight tier's gain depends on how failed sessions are handled, so the relationship between reviewer capability and the benefit of restriction is unclear.

\subsection{Reviewer Model Capability}

Within each restriction setting, F1 increases with reviewer tier: top-tier 32.1--32.3\% (above CCR's 28.6\%, not significantly), capable 26.8--29.0\% (approaching CCR with restriction), and lightweight 24.0--27.0\% (at or below the SR baseline of 27.1\%). The spread between the best and worst XMR conditions (8.3pp) is larger than the total spread among all same-model conditions (6.9pp), although ranges over different sets of conditions are not directly comparable effect sizes.

\subsection{Model Bias False Positives}\label{sec:biasfp}

Cross-model review introduces a new category of false positive: model bias errors, where the reviewer flags issues based on its own training data rather than actual defects. Classifying all XMR false positives via keyword-based analysis, 187 of 1,995 (9.4\%) were model bias; the most common subcategory (72.7\%, 136/187) involved flagging generator-specific features or tools as non-existent, which may reflect differences in training data between model families (the classification is keyword-based and was not validated). Information restriction partially mitigates this (lightweight 13\%$\to$10\%, capable 10\%$\to$8\%), and the top-tier model shows the lowest rate (9\% unrestricted, 8\% restricted).

The severity label a reviewer attaches is a weak guide to which findings to check first. Major or Critical labels were attached to 70.0\% of cross-model true positives and to 54.7\% of cross-model false positives (same-model: 54.5\% and 28.2\%; Appendix~\ref{app:severity}). Because a correct finding of a pre-existing defect counts as a false positive (Appendix~\ref{app:benchmark}), some of these false positives may be real defects; we have not checked them by hand.

\subsection{Case Studies}\label{sec:cases}

We ran five case studies outside software review---agricultural anomaly detection
(B1), recession identification (B2), news fact-checking (B3), a meta-review of this
paper's own analysis code (B4), and a security audit (B5). Because their evidential
value varies sharply, we summarize them here and report them in full in
Appendix~\ref{app:case}.

Only B3 (news fact-checking) is informative for our hypotheses: the top-tier
cross-model reviewer again has the highest F1, and the union of CCR and XMR-GPT findings reaches 86.0\% recall against
76.7\% for XMR-GPT alone (CCR 55.8\%), with a Jaccard of 54.1\%; each condition's recall here pools its three runs, so the union pools six reviews against three and is descriptive, not a matched comparison. B1 and B2 reached
100\% accuracy under every condition; a ceiling leaves no room to discriminate
between conditions, so we draw no comparative conclusion from them and report them
only as evidence that the task framing transfers. B4 and B5 lack independent ground
truth and are qualitative demonstrations, not tests. We therefore rest the
cross-domain claim on B3 alone, with the caveat that 12 articles from a single source
is a small sample.

\section{Discussion}

\subsection{Why Different Models Find Different Things}

The partial non-overlap is consistent with several explanations, though our design does not establish causal mechanisms: training-data divergence (different corpora, different standards), different capability profiles (the category reversal is consistent with code review benefiting from the generator model's own programming knowledge, document review from diverse perspectives), and model-specific knowledge gaps (cross-model reviewers also flag non-errors from their own gaps; Section~\ref{sec:biasfp}).

\citet{kim2025correlated} find strongly correlated errors on closed-ended benchmarks; our unions also fall below the independence reference (Section~\ref{sec:union}), so the partial non-overlap we observe should not be read as independent errors. Perspective diversification (B5) and three-model unions are discussed in Appendix~\ref{app:beyond}.

\subsection{Practical Guidelines}

Based on our findings, we offer the following tentative guidance, limited to this sample of artifacts and reviewers. For maximum coverage, combine same-model and cross-model review (56.7\% vs.\ 42.7\% for two same-model reviews at the same number of review calls; not significantly above two top-tier cross-model reviews). For a single review, top-tier cross-model review scored highest (F1 32.1--32.3\%), but not significantly above CCR (28.6\%). Prioritize same-model review for code-heavy projects (F1 40.7\% vs.\ 37.2\% on code) and cross-model review for documentation/consistency (F1 42.1\% vs.\ 24.5\% on documents); avoid cross-model review with lightweight models (F1 24.0--27.0\% vs.\ CCR 28.6\%).

\section{Conclusion}

We presented a four-dimensional framework for independence in LLM verification and new experiments on model independence. Across 900 sessions, CCR leads on code and XMR on documents, and one CCR plus one XMR review covers more ground-truth errors than two CCR reviews (56.7\% vs.\ 42.7\%). Most other comparisons are descriptive; in our primary test, only SR2 differs significantly from CCR. A factorial design that varies generator, reviewer model, and context independently, and evaluation on established benchmarks such as FEVER \citep{thorne2018fever}, AVeriTeC \citep{schlichtkrull2023averitec}, CUAD \citep{hendrycks2021cuad}, or LegalBench \citep{guha2023legalbench}, would test these patterns more directly.

\section*{Limitations}

\textbf{Design.} All artifacts came from a single generator, and each capability tier is represented by one reviewer model (three models from two developers), so reviewer capability and model identity are confounded. All cross-model conditions also include context independence, so H4 (an interaction between context and model independence) cannot be identified; a factorial design is left to future work. CCR and SR differ both in context and in whether the reviewer sees the original request, so their comparison does not isolate context independence. The GPT-5.4 reviews ran through the Codex CLI agent, which can read files in its working directory; we have no record of which directory that was. No GPT-5.4 finding mentions a ground-truth identifier or file, but that absence does not rule out access.

\textbf{Data and measurement.} The ground truth consists of planted errors written at injection time and was not independently annotated, and the matcher was not audited by hand. The same-model baselines reuse data collected before the cross-model conditions; their records cannot be checked against raw model output, and we treat them as unverified rather than clean (Section~\ref{sec:provenance}). The union comparison is retrospective: CCR runs differ widely (56, 37, and 29 matched errors), the headline pairing is among the higher of the reused run combinations, and whether the matcher treats the reviewers' output styles equally is unmeasured. The per-error-type and per-category comparisons are descriptive and were not tested for significance. Token counts were recorded only for the four Gemini conditions, so we cannot report cost per review or per detected error across conditions; the two-review-call comparison matches calls, not compute.

\textbf{Language.} All artifacts and ground-truth annotations are in Korean, and review output language was not fixed across conditions: in run~1, 20 of 30 CCR reviews are in English while the other conditions' reviews are mostly in Korean. Because the matcher's keyword component compares findings with Korean ground-truth text (Appendix~\ref{app:benchmark}), it may under-credit English-language reviews. Other languages are untested.

\textbf{Generalization.} The case studies are small: B1 and B2 showed ceiling effects, B3 has 12 articles from a single source, and in B5 one model returned no findings for one target. Our results are a snapshot of early-2026 models. A partial external-data check on ReaLMistake detector outputs (Appendix~\ref{app:realmistake}) neither replicates nor contradicts our main comparison.

\section*{Ethical Considerations}

Improved verification can catch errors in AI-generated code and analysis before they cause harm, but over-reliance is a risk: even our union approach matches only 56.7\% of the planted errors under our automated matcher. We emphasize that our framework is intended to \textit{augment}, not replace, human review; the undetected planted errors and the model-bias false positives (Section~\ref{sec:biasfp}) underscore the continued importance of human judgment for high-stakes applications.

\section*{Use of AI Assistants}

AI tools were used throughout this work, and we report their use as part of the method. The artifacts, the injected errors, and the same-model baseline reviews were produced by Claude Opus~4.6 through the Claude Code agent tool; the cross-model reviews were produced by the listed reviewer models, Gemini through its API and GPT-5.4 through the Codex CLI agent. Claude Code was also used to write the analysis and audit scripts, to recompute every reported number from the per-session records, to check the manuscript against the raw data (which led to the corrections described in Section~\ref{sec:provenance}), and to draft and revise substantial parts of the text under the author's direction. Every reference was checked for existence against bibliographic databases, and unresolved entries were checked by hand. The author reviewed the final text and takes full responsibility for all of its content.

\section*{Acknowledgments}

We thank Evgenii Arsentev for a public review of version~1 (PREreview, \url{https://doi.org/10.5281/zenodo.23132634}), which prompted the additions in version~2.

\section*{Data and Code Availability}

The per-session review records, the 30 artifacts with their 150 ground-truth defect annotations, and the matching, aggregation, and audit scripts are available from the author on request. They include the per-session sets of detected ground-truth identifiers, which are what the union and Jaccard figures in Section~\ref{sec:union} are computed from, so every set-level number in this paper can be recomputed independently.

\appendix

\section{Benchmark, Ground Truth, and Matching}
\label{app:benchmark}

\textbf{Artifacts.} The generator produced 30 artifacts from fixed prompts, all in Korean: ten Python functions or classes (C1--C10; e.g., CSV statistics, business-day calculation, an LRU cache), ten technical tutorials (D1--D10; e.g., Docker Compose, a FastAPI CRUD service, JWT authentication), and ten presentation scripts (S1--S10; talks on AI tooling, open data, and automation).

\textbf{Error injection and ground truth.} Into each artifact, exactly five errors were injected, one of each type (FACT, CONS, CTXT, RCVR, MISS), giving 150 errors: 30 Critical, 60 Major, and 60 Minor. According to our run logs, injection was performed by the generator model in dedicated injection sessions. Each ground-truth record stores the location, the original and injected text, the type, the severity, a description, and the correct answer. The records were written at injection time; we did not run an independent annotation study, so the ground truth is a set of planted errors rather than an exhaustive list of all defects in each artifact.

\textbf{Matching.} Each finding is scored against every ground-truth error of its artifact and assigned to the highest-scoring one if the score is at least 2.0: +3 if a line number in the finding's location equals one in the error's location, otherwise +1 if any are within five lines; plus $4\times$ the fraction of ground-truth keywords (tokens of three or more characters from the description, injected text, and correct answer) that appear in the finding's description; plus 1 if the finding's error type matches. Within one review, each ground-truth error is credited at most once, in the order findings are listed. Unmatched findings are counted as false positives, so precision and F1 are measured against the planted errors only; a finding that correctly identifies a pre-existing defect also counts as a false positive. A location match alone (3 points) clears the threshold, so the matcher can credit a finding that does not identify the error; we did not audit match accuracy by hand. The keyword component is language-sensitive (see Limitations).

\textbf{Statistics.} Per-condition means in Table~\ref{tab:main_results} average the analysed sessions ($n$ in the table). The primary test compares per-artifact means over the available runs against CCR (Wilcoxon signed-rank, 30 pairs, Holm over nine comparisons); run-1 paired $t$-tests are shown for comparison with the original CCR report. Precision, recall, and F1 are computed per session and then averaged, so the reported F1 is not the harmonic mean of the reported precision and recall. Per-type recall uses each session's single planted error of that type, so no session has an empty denominator.

\section{Review Prompts}
\label{app:prompts}

The prompts were written in Korean; we give English translations. The Korean originals are in the experiment code, and the information-restricted prompt is specified as the same string for CCR and the three -R conditions. The GPT-5.4 and Gemini~2.5 Pro prompts are as sent. The same-model baselines were run through the Claude Code agent tool (Section~\ref{sec:provenance}); the prompts below are those specified in the main experiment script, and the prompts those sessions actually received were not recorded. Separate scripts written for the excluded SR run~3 use a different self-review wording. The surviving Gemini script was edited from the Flash version to the Pro version; the Flash version was not preserved, so for the two Flash conditions the same prompt string is inferred rather than checked. All texts below are user-turn text only; the Codex CLI and the Claude Code agent tool add their own system instructions.

\textbf{Artifact only (CCR, XMR-Ge-R, XMR-GePro-R, XMR-GPT-R; also the review turn of SR, sent after the generation conversation).}
\begin{quote}\small
Verify the following [code / technical document / presentation script]. If there are errors, find them.
[artifact]
Specifically: 1. Factual accuracy: are numbers, names, and technical claims correct? 2. Internal consistency: are there contradictions? 3. Contextual fitness: are there problems in a real environment? 4. Recipient's perspective: is there anything a reader could misunderstand? 5. Completeness: is anything missing?
For each item, give the specific location and the grounds.
Response format (must be this JSON format): a list of findings, each with location (line number or paragraph/section), description, type (FACT|CONS|CTXT|RCVR|MISS), severity (Critical|Major|Minor), and suggestion; and a one-to-two-sentence summary.
\end{quote}

\textbf{Request and artifact (SA, XMR-Ge, XMR-GePro, XMR-GPT).}
\begin{quote}\small
The following is a prompt a user gave to an AI and its output. Verify whether the output has errors.
Original request: [generation prompt]. Generated output: [artifact].
The five checks, the instruction to give locations and grounds, and the JSON format follow as above (for SA, the format block is worded slightly more briefly).
\end{quote}

\textbf{Second pass (SR2, after a first artifact-only review in the same conversation).}
\begin{quote}\small
Thanks. But something may have been missed, so please verify it again from the beginning, focusing especially on parts you did not point out before. The five checks and the instruction to give locations and grounds follow as above, with the JSON format given in abbreviated form.
\end{quote}

\section{Severity Labels of True and False Positives}
\label{app:severity}

\begin{table}[h]
\centering
\small
\begin{tabular}{llrrrr}
\toprule
\textbf{Group} & & $n$ & \textbf{Crit.} & \textbf{Major} & \textbf{Minor} \\
\midrule
Same-model  & TP & 374   & 22.5 & 32.1 & 45.5 \\
            & FP & 1,106 & 6.1  & 22.2 & 70.0 \\
Cross-model & TP & 771   & 19.7 & 50.3 & 30.0 \\
            & FP & 1,988 & 12.9 & 41.8 & 45.3 \\
\bottomrule
\end{tabular}
\caption{Severity labels attached by reviewers (\% of findings). Analysed sessions as in Table~\ref{tab:main_results}, minus 26 sessions (24 SR, 2 XMR-Ge-R) whose findings were recovered from raw output and cannot be aligned with their labels. Because 24 of the 26 are SR sessions, SR contributes 36 of its 60 analysed sessions to the same-model group, and the cross-model false positives number 1,988 rather than the 1,995 of Section~\ref{sec:biasfp}. Twenty same-model false positives carry other labels and are omitted from the percentages' numerators; sums may differ by 0.1 from rounding. Descriptive; not tested.}
\label{tab:severity}
\end{table}

\section{External-Data Check on ReaLMistake}
\label{app:realmistake}

As a partial external-data check, we recomputed the two-review-call comparison on the public detector outputs of ReaLMistake \citep{kamoi2024evaluating}, without new model calls. There, a ``same-model review'' is the responder model detecting errors in its own responses, the two same-model reviews differ only in prompt wording (temperature 0), and the cross-model partner is Claude~3 Opus. The analysis plan was committed to our version-controlled repository before the results were computed (available on request). Pooled over three tasks and two responder models (649 erroneous and 251 correct responses), union recall of the mixed pair was not detectably different from two same-model reviews ($+0.5$pp, 95\% CI [$-1.2$, $2.2$]; item bootstrap stratified by cell). In a post hoc split by responder, mixing raised recall for GPT-4 responses and lowered it for Llama-2-70B responses, and in every cell the false-positive rate moved in the same direction as recall, so the split is consistent with a shift in operating point rather than with added coverage alone. Two Claude~3 Opus reviews did not consistently trail the mixed pair, so the check does not separate model difference from reviewer capability. Because the unit (response-level binary labels), the source of variation between same-model reviews, the domain, and the detector generation (early 2024) all differ from our setting, this check neither replicates nor contradicts our main result.

\section{Additional Figure}
\label{app:fig}

This figure presents the same data as Table~\ref{tab:main_results} in graphical form.

\section{Values With Failed Sessions Scored as Zero}
\label{app:itt}

Table~\ref{tab:itt} compares the main per-condition F1 values with and without the exclusions of Section~\ref{sec:provenance}. ``All sessions'' keeps SR run~3 and scores the 14 failed cross-model sessions as reviews with no findings, which is how the values are computed without the exclusions.

Run-level F1 for the same-model conditions (30 artifacts each; runs 1/2/3) is: CCR 37.2/25.0/23.7, SR 25.7/28.6/(19.4, excluded), SA 23.8/20.8/26.9, and SR2 18.7/22.5/23.9. CCR ranks first among them in run~1 only, which is the run the original CCR report tested.

\begin{table}[h]
\centering
\small
\begin{tabular}{lccc}
\toprule
\textbf{Condition} & \textbf{All sessions} & \textbf{Excluded} & \textbf{$n$ excl.} \\
\midrule
XMR-GPT     & 30.9 & 32.3 & 86 \\
XMR-GPT-R   & 31.8 & 32.1 & 89 \\
XMR-GePro-R & 29.0 & 29.0 & 90 \\
CCR         & 28.6 & 28.6 & 90 \\
SR          & 24.5 & 27.1 & 60 \\
XMR-Ge-R    & 24.6 & 27.0 & 82 \\
XMR-GePro   & 26.8 & 26.8 & 90 \\
XMR-Ge      & 23.7 & 24.0 & 89 \\
SA          & 23.8 & 23.8 & 90 \\
SR2         & 21.7 & 21.7 & 90 \\
\bottomrule
\end{tabular}
\caption{F1 (\%) with all 900 sessions and after the exclusions. With all sessions, the primary test against CCR (Holm over nine comparisons) finds only SR2 different ($p=0.011$); SR's unadjusted $p$ is 0.021 (0.26 after excluding its run~3), and XMR-GPT vs.\ XMR-Ge has unadjusted $p=0.071$ (0.020 after excluding failures).}
\label{tab:itt}
\end{table}

\section{Additional Discussion}

\subsection{Beyond Two Models: Diminishing Returns}\label{app:beyond}

Our primary analysis is pairwise, but adding a third reviewer model yields diminishing returns. In Experiment~A, the three-model union (CCR $\cup$ XMR-GPT-R $\cup$ XMR-GePro-R) reaches 58.0\% recall (run~1), only 1.3pp over the two-model union (56.7\%). In this setting, a third reviewer adds little.

\subsection{Perspective Diversification}

Our B5 case study further suggests that \textit{prompt perspective} diversification produces partly non-overlapping findings even within one model, though model diversification yields lower overlap (Table~\ref{tab:perspective-jaccard}); the architect perspective yields the lowest inter-model overlap, an untested hint that design-level assessment may differ most across models. One model's anomalous zero-finding behavior cautions, however, that not all models respond uniformly to perspective variation.

\section{Case Studies in Full}
\label{app:case}

\subsection{Setup}

To assess generalizability beyond code review, we conduct five case studies (B5's setup is given with its results):

\textbf{B1: Agricultural price anomaly detection.} From national open-data agricultural auction records (2022--2026, 1,155 trading days, 15 commodities), we flagged 15 candidate anomalies via Z-score ($|z|>3$) and IQR intersection and added 3 normal-range controls; one annotator labeled the candidates by hand (14 true anomalies, 1 uncertain). Both same-model and cross-model reviewers analyzed each.

\textbf{B2: Economic recession identification.} Using macroeconomic time series (GDP growth, unemployment, yield-curve spread, Sahm Rule indicator; 1960--2025), we designed six tasks requiring recession-period identification, with official government-designated dates as ground truth.

\textbf{B3: News fact-checking.} Using 12 PolitiFact-verified articles across six topic areas (ratings: Half True 4, Mostly False 3, False 3, Pants on Fire 2), the generator analyzed each article's claims; three reviewer conditions (CCR, XMR-GPT, XMR-Ge) each reviewed all 12 three times (108 reviews total), against 43 ground-truth errors from PolitiFact's analyses; the article list and error annotations are available with the other records.

\textbf{B4: Meta case study---reviewing this paper's code.} We applied same-model review (fresh session) and cross-model review to the visualization code written for this paper's figures, providing a qualitative illustration of the non-overlap measured quantitatively in Experiment~A.

\subsection{Results}

\textbf{B1: Agricultural anomaly detection.} All conditions classified all 17 cases with a definite label correctly (14 true anomalies and 3 normal controls); the one case the annotator marked uncertain was treated the same way by every condition and is not scored. The anomalies involved extreme price spikes ($z > 3$, increases of 42--138\%) unambiguous to any capable model.

\textbf{B2: Economic recession identification.} Similarly, all conditions reached 100\% accuracy across 6 tasks, correctly identifying all officially designated recession periods; timing error was 0 months (same-model) and 0.2 months (cross-model).

\textbf{B3: News fact-checking.} To test generalization beyond software artifacts, we conducted a fact-checking case study with 12 PolitiFact-verified articles (43 ground-truth errors, 108 reviews). The reviewer conditions achieved F1 of 20.7\% (CCR), 24.4\% (XMR-GPT), and 23.8\% (XMR-Ge), with XMR-GPT highest in recall (76.7\%, vs.\ 55.8\% for CCR and 51.2\% for XMR-Ge); precision was low (12.7--15.5\%) in all conditions. The union recall of CCR and XMR-GPT reached 86.0\% (Jaccard 54.1\%; 3-way union 88.4\%). Because each condition's recall already pools its three runs, this union pools six reviews against three and is not a matched-call comparison; we report it descriptively.

The fact-checking results resemble Experiment~A's key patterns in a non-software domain: in both settings the top-tier cross-model reviewer has the highest individual F1 (Experiment~A overall 32.3\% vs.\ CCR 28.6\%, not significant; fact-check 24.4\% vs.\ 20.7\%, untested), and neither shows ceiling effects. The higher Jaccard in fact-checking (54.1\% vs.\ 41.2\%) may indicate that some errors are detectable across models; the fact-checking union (pooled over runs) is not comparable with Experiment~A's call-matched union, and sample composition, detection rates, and matching also differ between the two settings, and B3 has only 12 articles.

\textbf{Interpretation of B1, B2, and B3.} B1 and B2 do not discriminate between conditions. One untested possibility is that differences appear only on judgment-intensive tasks; we did not manipulate or measure task difficulty. In these two cases, where the signal was strong (extreme price spikes, well-documented recessions), every condition we tested succeeded. The differences measured in Experiment~A and B3, where review requires nuanced judgment about correctness, consistency, and completeness, may represent the regime where independence dimensions matter. B3 suggests that this regime is not specific to software artifacts: fact-checking of news claims produces similar qualitative patterns.

\textbf{B4: Meta case study.} Same-model and cross-model review independently examined our visualization code (2 Python files, $\sim$600 lines): same-model review found 15 issues (2 P1, 8 P2, 5 P3; reviewer-assigned priority levels, P1 highest) and the cross-model reviewer 6 (3 P1, 3 P2), with 4 in common (Jaccard = 4/17 = 23.5\%). In this case, the same-model findings were mostly subtle logic inconsistencies and convention violations, while the cross-model findings included edge cases, among them a crash-inducing path that same-model review did not report; we did not apply an independent judgement procedure to these findings.

In a second round reviewing this paper's draft text, same-model review found 27 issues and the cross-model reviewer 15; we no longer have the cluster-level records for this round and do not report an overlap figure for it. In this round, the reported same-model findings concerned cross-referencing and framework consistency, and the cross-model findings overclaiming and source traceability. The code-review overlap (23.5\%), below the 41.2\% in Experiment~A, suggests overlap may vary with task characteristics, though B4 is a qualitative demonstration rather than a statistical test.

\textbf{B5: Security audit perspective diversification.} To test whether varying the \textit{perspective} (rather than the model) produces non-overlapping findings even within the same model, we conducted a security audit on two C libraries (wren, 8.9K lines; cgltf, 3.1K lines) reviewed by three frontier models from distinct families under three perspectives: attacker (exploit-focused), defender (mitigation-focused), and architect (design-level). Findings were matched by file, function, and CWE category.

\begin{table}[t]
\centering
\small
\resizebox{\columnwidth}{!}{%
\begin{tabular}{lcc}
\toprule
\textbf{Comparison axis} & \textbf{Mean Jaccard} & \textbf{Range} \\
\midrule
Inter-perspective (same model) & 0.228 & 0.143--0.331 \\
Inter-model (same perspective) & 0.152 & 0.057--0.259 \\
\bottomrule
\end{tabular}}
\caption{Perspective vs.\ model diversification Jaccard overlap (lower means less overlap).}
\label{tab:perspective-jaccard}
\end{table}

Both diversification axes produce low overlap (Table~\ref{tab:perspective-jaccard}), below the 41.2\% CCR--XMR overlap observed in Experiment~A, indicating partly non-overlapping findings; low overlap alone does not show that the additional findings are correct. Mean inter-model overlap is about a third lower than inter-perspective overlap ($J = 0.152$ vs.\ $0.228$), consistent with changing the model producing more distinct findings than changing the prompt alone.

The architect (design-level) perspective produces the lowest inter-model Jaccard ($J = 0.075$; wren 0.082, cgltf 0.068) and the attacker perspective the highest (wren 0.208, cgltf 0.157). One untested reading is that exploit-oriented analysis follows more standardized patterns across models than design-level reasoning; differences in finding counts, output format, and matching could also produce this pattern. Version~1 also reported percentage gains in finding counts from combining perspectives or models; we could not re-derive their base counts and have removed them.

One anomaly limits generality: one model produced zero findings for the smaller target (cgltf, 3.1K lines) under the defender and architect perspectives while functioning normally otherwise; we did not determine whether this reflects an absence of defects, a detection failure, or an output failure. Whether perspective and model diversification are complementary is a hypothesis these data do not test: the findings were not validated as defects, the union over all nine model--perspective combinations was not measured, and two targets are a limited sample.

\bibliography{references}

\end{document}